\documentclass[a4paper,fleqn]{cas-sc}
\usepackage{lipsum}
\makeatletter
\def\ps@pprintTitle{%
	\let\@oddhead\@empty
	\let\@evenhead\@empty
	\def\@oddfoot{}%
	\let\@evenfoot\@oddfoot}
\makeatother
\ExplSyntaxOn \cs_gset:Npn \__first_footerline: { \group_begin: \small \sffamily \__short_authors: \group_end: } \ExplSyntaxOff
\usepackage[authoryear,longnamesfirst]{natbib}
\usepackage{bm}
\usepackage{amsmath,graphicx}

\usepackage[caption=false,font=footnotesize]{subfig}
\usepackage{amsthm}
\usepackage{array}
\usepackage{float}
\newcolumntype{P}[1]{>{\centering\arraybackslash}p{#1}}
\newcolumntype{M}[1]{>{\centering\arraybackslash}m{#1}}
\def\tsc#1{\csdef{#1}{\textsc{\lowercase{#1}}\xspace}}
\tsc{WGM}
\tsc{QE}
\tsc{EP}
\tsc{PMS}
\tsc{BEC}
\tsc{DE}

\begin{document}
	\let\WriteBookmarks\relax
	\def\floatpagepagefraction{1}
	\def\textpagefraction{.001}
	\shorttitle{Information Fusion: Scaling Subspace-Driven Approaches}
	\shortauthors{Sally Ghanem et~al.}
	
	\title [mode = title]{Information Fusion: Scaling Subspace-Driven Approaches}

	\author{Sally Ghanem}[orcid=0000-0003-0186-820X]
	\cormark[1]
	\ead{ssghanem@ncsu.edu}

	\author%
	{and Hamid Krim}

	\address{Electrical and Computer Engineering Department, North Carolina State University, Raleigh, North Carolina, 27695.}
	
	\cortext[cor1]{Corresponding author}
	
	\begin{abstract}
		In this work, we seek to exploit the deep structure of multi-modal data to robustly exploit the group subspace distribution  of the information using the Convolutional Neural Network (CNN) formalism. Upon unfolding  the set  of subspaces constituting  each  data modality, and learning their corresponding encoders, an optimized  integration of the generated inherent information is carried out to yield a characterization of various classes. Referred to as deep Multimodal Robust Group Subspace Clustering (DRoGSuRe), this approach is compared against the independently developed state-of-the-art approach named Deep Multimodal Subspace Clustering (DMSC). Experiments on different multimodal datasets show that our approach is competitive and more robust in the presence of noise.
	\end{abstract}

	\begin{keywords}
		Sparse learning \sep Computer vision \sep Unsupervised classification \sep  Subspace clustering \sep Multi-modal sensor data 
	\end{keywords}

	\maketitle

	\section{Introduction}
	Unsupervised learning is a very challenging topic in Machine Learning (ML), and involves the discovery of hidden patterns in data for inference with no prior given labels. Reliable clustering techniques will save time and effort required for classifying/labeling large datasets that might have thousands of observations. Multi-modal data, increasingly in need for complex application problems, have become more accessible with recent advances in sensor technology, and of pervasive use in practice. The plurality of sensing modalities in our applications of interest, provides diverse and complementary information, necessary to capture the salient characteristics of data and secure their unique signature. A principled combination of the information contained in the different sensors and at different scales is henceforth pursued to enhance understanding of the distinct structure of the various classes of data. The objective of this work is to develop a principled multi-modal framework for object clustering in an unsupervised learning scenario. We extract key class-distinct features-signatures from each data modality using a CNN encoder, and we subsequently non-linearly combine those features to generate a discriminative characteristic feature. In so doing, we work on the hypothesis that each data modality is approximated by a Union of low dimensional Subspaces which highlights underlying hidden features. The UoS structure is unveiled by pursuing sparse self-representation of the given data modality. The subsequent aggregation of the multi-modal subspace structures yields a jointly unified characteristic subspace for each class.
	\subsection{Related Work}
	
	Subspace clustering has been introduced as an efficient way for unfolding union of low-dimensional subspaces underlying high dimensional data. Subspace clustering has been extensively studied in computer vision due to the vast availability of visual data as in \cite{elhamifar2013sparse}, \cite{favaro2011closed}, \cite{li2015structured}, and \cite{bian2018bi}. This paradigm has broadly been adopted in many applications such as image segmentation \cite{yang2006robust}, image compression \cite{hong2006multiscale}, and object clustering \cite{ho2003clustering}.
	Uncovering the principles and laying out the fundamentals for multi-modal data has become an important topic in research in light of many applications in diverse fields including image fusion \cite{hellwich2000object}, target recognition \cite{korona1996model}, \cite{ghanem2018information}, \cite{ghanem2021latent}, \cite{wang2018fusing}, speaker recognition \cite{soong1988use}, and handwriting analysis \cite{xu1992methods}. Convolutional neural networks have been widely used on multi-modal data as in \cite{ngiam2011multimodal} \cite{ramachandram2017deep}, \cite{valada2016deep}, \cite{roheda2018cross}, and \cite{roheda2018decision}. A very similar approach for multi-modal subspace clustering was proposed in \cite{zhu2019multi}. However, we note and maintain that our formulation results in a different
	optimization problem, as the multi-modal sensing seeks to not only account for the private information which provides the complementarity of the sensors, but also the common and hidden information. This yields, naturally, to a different network structure than that of \cite{zhu2019multi} in a different application space. In addition, the robustness of fusing multi-modal sensor data each with its distinct intrinsic structure, is addressed along with a potential scaling for viability. A thorough comparison of our results to the state of the art multimodal fusion network \cite{abavisani2018deep} is carried out, with a demonstration of resilient fusion under a variety of limiting scenarios including limited sensing modalities (sensor failures).
	\subsection{Contributions}
	Building on the work of Deep Subspace Clustering (DSC) \cite{ji2017deep}, we propose a new and principled multi-modal fusion approach which accounts for a sensor' capacity to house private and unique information about some observed data as well as that information which is likely also captured and hence common to other sensors. This is accounted for  in our robust fusion formulation for multi-modal sensor data. Unveiling the complex UoS of multi-modal data also requires us to account for scaling in our proposed formulation and solution, which in turn invokes the learning of multiple/deep scale Convolutional Neural Networks (CNN). Our proposed Multi-modal fusion approach, by virtue of each sensor information structure (i.e. private plus shared) seeks to  enhance and robustify the subspace approximation of shared information for each of the sensors, thus yielding a parallel bank of UoS for each of the sensors. The robust  Deep structure effectively achieves scaling while securing structured representation for  unsupervised inference. We compare our approach to a well-known deep multimodal network \cite{abavisani2018deep} which was also based on \cite{ji2017deep}.\\
	
	In our proposed approach, we thus define the latent space in a way that safeguards the individual sensor private information which hence dedicates more degrees of freedom to each of the sensors. In contrast to the approach in \cite{abavisani2018deep}. In our evaluation, we use two recently released data sets each of which we partition into learning and validation subsets. The learned UoS structure for each of the data sets is then utilized to classify new observed data points, which illustrates the generalization power of the proposed approach. Different scenarios with corresponding additive noise to either the training set or the testing set, or both, were used to thoroughly investigate the robustness, and resilience of the clustering approach performance. Experimental results confirm a significant improvement for our Deep Robust Group Subspace Recovery network (DRoGSuRe) under numerous limiting scenarios and demonstrate robustness under these conditions.\\ 
	
	The balance of the paper is organized as follows, in Section 2, we provide the problem formulation, background along with the derivation for our proposed approach, Deep Robust Group Subspace Recovery (DRoGSuRe). In Section 3, we describe the attributes of the proposed approach and contrast it to Deep Multimodal Subspace Clustering algorithm (DMSC). In Section 4 and 5, we present a substantiative validation along with experimental results of our approach, while Section 6 provides concluding remarks.
	
	\section{Deep Robust Group Subspace Clustering}
	\subsection{Problem Formulation}
	We assume having a set of data observations, each represented as a $m-$dimensional vector  $\mathbf{x_k}(t) \in \mathbb{R}^m $, where \textit{k = 1, 2, ..., n}. Moreover, we consider having \textit{T} data modalities, indexed by \textit{t = 1, 2, 3, ..., T}. Each data modality can then be described as $\mathbf{X}(t)=[\mathbf{x}_1(t) \  \mathbf{x}_2(t) \ ... \ \mathbf{x}_n(t)]$. Our objective is to assign each set of data observations into clusters that can be efficiently represented by a low-dimensional subspace. This is equivalent to finding a partitioning \{$\mathbf{X}^1(t),\mathbf{X}^2(t), ...,\mathbf{X}^P(t)$\} of [$n$] observations, where $P$ is the total number of clusters underlying each data modality indexed by $p$. Furthermore, each linear subspace can be described as $\mathbf{S}^p(t) \subset \mathbb{R}^m$ with dim($\mathbf{S}^p(t)$) $ \ll m$.\\
	We will exploit the self-expressive property presented in \cite{elhamifar2013sparse} and \cite{bian2015bi}, which entails that each data observation $\mathbf{x}_i(t)$ can be represented as a linear combination of all features from the same subspace $S(\mathbf{x}_i(t))$ as follows,
	\begin{equation}
		\mathbf{x}_i(t)=\sum_{i\neq j, \mathbf{x}_j(t) \in S(\mathbf{x}_i(t))} w_{ij}(t) \mathbf{x}_j(t) 
	\end{equation}
	If we stack all the data points  $\mathbf{x}_i(t)$ into columns of the data matrix $\mathbf{X}(t)$, The self-expressive property can be written in a matrix form as follows, 
	\begin{equation}
		\mathbf{X}(t)=\mathbf{X}(t)\mathbf{W}(t) s.t. \mathbf{W_{ii}}=0.
	\end{equation}
	The important information about the relations among data samples is then recorded in the self-representation coefficient matrix $\mathbf{W}(t)$. Under a suitable arrangement/permutation of the data realizations, the sparse coefficient matrix $\mathbf{W}(t)$ is an $n \times n$ block-diagonal matrix with zero diagonals provided that each sample is represented by other samples only from the same subspace. More precisely, $W_{ij}(t)=0$ whenever the indexes $i,j$ correspond to samples from different subspaces. As a result, the majority of the elements in $W$ are equal to zero. A diagram showing our algorithm is depicted in Figure 1.\\
	Our algorithm consists of three main stages; the first stage is the encoder which encodes the input modalities into a latent space. The encoder consists of $T$ parallel CNN networks,  where $T$ is the number of data modalities. Each modality data is fed into one network, and the output of each network represents the modality data projection into its corresponding hidden/latent space. The second component of the auto encoder is $T$ self-expressive layers, the goal of which is to enforce the self-expressive property among the data observations of each data modality. Each self-expressive layer is a fully connected layer which independently operates on the output of each encoder. The last stage is the decoder which reconstructs input data from the self-expressive layers' output. The objective function sought through this approximation network is reflected in Equation (5). The group sparsity introduced in \cite{ghanem2020robust} requires the minimization of the group norm of matrices W(t), which in turn, entails a smaller angle between the different spaces across all modalities, thus promoting  the goal of obtaining a common latent space. Note that minimizing group norm provides as well a group sparse solution along data modalities. If we in addition, constrain the coefficient matrices corresponding to each data modality to commute, therefore, we ensure their sharing the same eigen vectors. The idea of commutation has been used in \cite{roheda2020robust}, \cite{roheda2020commuting}, and \cite{roheda2019event}. We define  $ \mathbf{\Omega} = \{ \mathbf{W} (t) \}_{t=1} ^T$, where $ \mathbf{W} (t)=[w_{kj}(t)]_{k,j}$ and the group l-norm $\parallel \mathbf{\Omega} \parallel_{1,2}$ as:
	\begin{equation}
		\parallel \mathbf{\Omega} \parallel_{1,2}=\sum_{k,j}\sqrt{\sum_{t=1}^T w_{k,j}^2(t)}.
	\end{equation}
	We also define $[W(t_1),W(t_2)]$ as,
	\begin{equation}
		[W(t_1),W(t_2)]=W(t_1)W(t_2)-W(t_2)W(t_1)=0.
	\end{equation}
	The loss function is then rewritten as,
	\begin{equation}
		\begin{split}
			\min\limits_{\mathbf{W(t)} \mid w_{kk}(t)=0}  &\sum_{t_1,t_2=1}^{T} \parallel [W(t_1),W(t_2)] \parallel^2+ \parallel \Omega \parallel_{1,2}+
			\frac{\gamma}{2} \sum_{t=1}^{T} \parallel X(t)-X_r(t) \parallel_F^2 +\rho \sum_{t=1}^T \parallel \mathbf{W}(t) \parallel_1+\\
			&\frac{\mu}{2} \sum_{t=1}^{T} \parallel L(t)-L(t)W(t) \parallel_F^2.
		\end{split}  
	\end{equation}
	
	Where $X_r(t)$ represent the reconstructed data corresponding to modality $t$, and $L(t)$ is the output of the $t^{th}$ encoder with input $X(t)$. $W(t)$ is the sparse weight function that ties the data observation for modality $t$. Solving DRoGSuRe in Tensorflow and using the adaptive momentum based gradient descent method (ADAM) \cite{kingma2014adam} results in minimizing the loss function. For each data modality, the weights of the encoder, the self-expressive layer and the decoder are individually calculated, however, fine-tuning the weights is based on the loss function, which is a function of the group norm and the pairwise product difference between sparse coefficient matrices. $\| \|_1$ denotes the $l_1$ norm, i.e. the sum of absolute values of the argument. 
	The Lagrangian objective functional may be rewritten as,
	\begin{equation}
		\begin{split}
			L(\mathbf{\Omega}, \mathbf{W}(t), \mathbf{Y}(t), \mu)=&\sum_{t_1,t_2=1,t_1\neq t_2}^{T} \parallel [W(t_1),W(t_2)] \parallel^2+\parallel \Omega \parallel_{1,2}+ \rho \sum_{t=1}^T \parallel \mathbf{W}(t) \parallel_1 +\frac{\gamma}{2} \sum_{t=1}^{T} \parallel X(t)-X_r(t) \parallel_F^2+\\ &\sum_{t=1}^T \frac{\mu}{2}\parallel\mathbf{L}(t)\mathbf{W}(t)-\mathbf{L}(t)\parallel_F^2+\sum_{t=1}^T<\mathbf{L}(t)\mathbf{W}(t)-\mathbf{L}(t),\mathbf{Y}(t)>.
		\end{split}
	\end{equation}
	Assume $\mathbf{\hat{W}}(t)=\mathbf{I}-\mathbf{W}(t)$, we update $\mathbf{W}(t)$ as follows,
	\begin{equation}
		\begin{split}
			\mathbf{W}_{k+1}(t)=&\arg \min\limits_{\mathbf{W}(t)} \sum_{t,m=1, t\neq m}^{T} \parallel [W(t),W(m)] \parallel^2+\parallel \mathbf{\Omega} \parallel_{1,2}+\rho \parallel \mathbf{W}(t)\parallel_1 + <\mathbf{L}_{k+1}(t)\mathbf{W}(t)-\mathbf{L}_{k+1}(t), \mathbf{Y}_k(t)>+\\& \frac{\mu_k}{2}\parallel\mathbf{L}_{k+1}(t)\mathbf{W}(t)-\mathbf{L}_{k+1}(t)\parallel_F^2
		\end{split}
	\end{equation}

	Similar to \cite{bian2018bi}, we utilize linearized ADMM \cite{lin2011linearized} to approximate the minimum of Eqn.(7) since the algorithmic solution is complicated and yields a non-convex optimization functional. It has been shown that linearized ADMM is very effective for $l_1$ minimization problems and the augmented Lagrange multiplier (ALM) method can take care of the non-convexity of the problem \cite{rockafellar1974augmented} \cite{luenberger1984linear}. Therefore, utilizing an appropriate augmented Lagrange multiplier $\mu_k$, we can compute the global optimizer by solving the dual problem. The solution to Eqn. (7) can be approximated, using linearized soft-thresholding, as follows,
	\begin{equation}
		\begin{split}
			\mathbf{W}_k^+(t)=& \mathrm{prox}_\frac{\rho}{\mu\eta_1}(\sum_{m,t=1,m \neq t}^T \{(\mathbf{W}_k(t)\mathbf{W}_k(m)-\mathbf{W}_k(m)\mathbf{W}_k(t))\mathbf{W}_k(m)^T+ \mathbf{W}_k(m)(\mathbf{W}_k(t)\mathbf{W}_k(m)-\mathbf{W}_k(m)\mathbf{W}_k(t)) \}+\\& \mathbf{W}_k(t)+ \frac{\mathbf{L}^T_{k+1}(\mathbf{L}_{k+1} \hat{W}_k(t)-\frac{\mathbf{Y}_k(t)}{\mu_k})}{\eta_1})
		\end{split}
	\end{equation}
	
	\begin{equation}
		\mathbf{W}_{k+1}(t)=\gamma_\frac{\rho}{\mu\eta_1}(\mathbf{W}_k^+(t))
	\end{equation}
	where $\eta_1 \geq \parallel L \parallel ^2_2$. We alternatively update $\mathbf{L}(t)$ as,
	\begin{equation}
		\mathbf{L}_{k+1}(t)=\mathbf{L}_k(t)+\mu_k(\mathbf{L}_{k}(t)\hat{\mathbf{W}}_{k+1}(t)-\frac{\mathbf{Y}_k(t)}{\mu_k})\hat{\mathbf{W}}^T_{k+1}(t)
	\end{equation}
	where $\mathrm{prox}_{\beta}(A_{i,j}(t))=A_{i,j}(t)*\max\{ (\sqrt{\sum_{t=1}^T A_{i,j}(t)^2} -\beta),0     \}/\sqrt{\sum_{t=1}^T A_{i,j}(t)^2}$ and $\gamma_\tau(B_{i,j})= sign(B_{i,j})*\max \{ (\mid B_{i,j}\mid-\tau),0 \}$. The Lagrange multipliers are updated as follows,
	
	\begin{equation}
		\mathbf{Y}_{k+1}(t)=\mathbf{Y}_k(t)+\mu_k(\mathbf{L}_{k+1}(t)\mathbf{W}_{k+1}(t)-\mathbf{L}_{k+1}(t))
	\end{equation}
	\begin{equation}
		\mu_{k+1}=\epsilon \mu_k.
	\end{equation}

	After computing the gradient of the loss function, the weights of each multi-layer network, that corresponds to one modality, are updated while other modalities' networks are fixed. In other words, after constructing the data during the forward pass, the loss function determines the updates that back-propagates through each layer. The encoder of the first modality is updated, afterwards, the self-expressive layer of that modality gets updated and finally the decoder. Since the weights corresponding to each modality are dependent on other modalities, we update each part of the network corresponding to each modality with the assumption that all other networks' components corresponding to other modalities are fixed. The resulting sparse coefficient matrices $\mathbf{W}(t)$'s, for $t= 1,2,...,T$ are then integrated as follows,
	\begin{equation}
		\mathbf{W}_{Total}=\sum_{t=1}^T \mathbf{W}(t).
	\end{equation} 
	Integrating the sparse coefficient matrices helps reinforcing the relation between data points that exist in all data modalities, thus establishing a cross-sensor consistency. Furthermore, adding the sparse coefficient matrices reduces the noise variance introduced by the outliers. A similar approach was introduced in \cite{taylor2016enhanced} for Social Networks community detection, where an aggregation of multi-layer adjacency matrices was proved to provide a better Signal to Noise ratio, and ultimately better performance. To proceed with distinguishing the various classes in  an unsupervised manner, we construct the affinity matrix as follows, 
	\begin{equation}
		\mathbf{A}=\mathbf{W}_{Total}+\mathbf{W}^T_{Total}.
	\end{equation}
	where $A \in \mathbb{R}^{n \times n} $. We subsequently use the spectral clustering method \cite{ng2002spectral} to retrieve the clusters in the data using the above affinity matrix as input.
	\begin{figure}[htb]
		\centering
		\includegraphics[width=13cm,height=5cm]{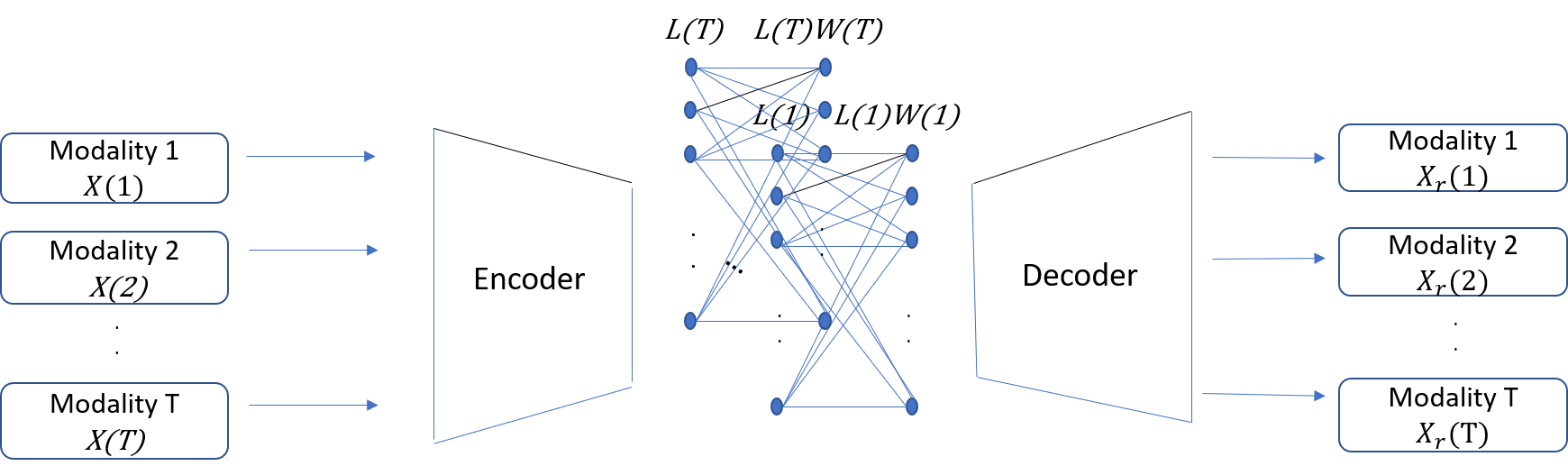}

		\caption{Deep Robust Group Subspace Clustering}.
	\end{figure}
	\subsection{Theoretical Discussion}
	In order to justify the multiple banks of self-expressive layers, we assume that each modality $\mathbf{X}(t)$ may be expressed as a private information contribution $X_p(t)$ and a shared information $X_s(t)$ such that,
	\begin{equation}
		X(t)=X_s(t)+X_p(t)
	\end{equation}
	The shared information can be represented as follows,
	\begin{equation}
		X_s(t)=\sum_{t=1}^T f(W(t)(\Pi_sX(t))),
	\end{equation}
	where $\Pi_s= \bigcap_{t=1\cdots T} \Pi_s^t$. $X_s(t)$ and $X_p(t)$ are distinct and will hence lie in different subspaces, which will hence be mapped to different components in $\mathbf{W}(t)$. Similarly for the subspaces spanned by $X_p(t_i)$ and $X_p(t_j)$,  $i\neq j$, the corresponding components of $\mathbf{W}(t_i)$ and $\mathbf{W}(t_j)$ will almost surely  not coincide. On the other hand, the components of  $\mathbf{W}(t_i)$ and $\mathbf{W}(t_j)$ corresponding to $X_s(t_i)$ and $X_s(t_j)$ will almost surely coincide, thus justifying the construction of a layered  $\mathbf{W}_{Total}$, and thereby improving the SNR. In addition, the decoder will help protect and maintain the private information corresponding to each modality $X_p(t)$ by ensuring that data can be reconstructed again from the latent space with minimal loss. In the following, we will elaborate more on how aggregating affinity matrices should impact the overall clustering performance. The idea of aggregating affinity matrices is not new, in fact, it has been used extensively in clustering and community detection field. For example, in \cite{gabriel2010eigenvector}, the authors proposed a method that combines the self-similarity matrices of the eigenvectors after applying a Singular Value Decomposition on clusters. In \cite{dong2013clustering}, they proposed merging the information provided by the multiple modalities by combining the characteristics of individual graph layers using tools from subspace analysis on a Grassmann manifold. In \cite{chen2017multilayer}, they propose a multilayer spectral graph clustering (SGC) framework that performs convex layer aggregation.
	
	\subsubsection*{Proposition} The persistent differential scaling of $\mathbf{m}$-modal Group Robust Subspace Clustering Fusion yields an order $\mathbf{m}$-improvement resilience over the singly differential scaling fusion
	
	The proof of the proposition can be found in Appendix A. We basically show that by perturbing one or more data modalities, our proposed approach introduces less error to the overall affinity matrix as compared to DMSC. Hence, preserving the performance and yielding a graceful degradation of the clustering accuracy as an increasing number of modalities get corrupted by noise.
	
	\section{Affinity Fusion Deep Multimodal Subspace Clustering}
	For completeness, we provide a brief overview of the Deep Multimodal Subspace Clustering algorithm which was proposed in \cite{abavisani2018deep}. As noted earlier for DRoGSuRe and similarly for Affinity Fusion Deep Multimodal Subspace clustering (AFDMSC), the network is composed of three main parts: a multimodal encoder, a self-expressive layer, and a multimodal decoder. The output of the encoder contributes to a common latent space for all modalities. The self-expressiveness property applied through a fully connected layer between the encoder and the decoder results in one common set of weights for all the data sensing modalities. This marks a divergence in defining the latent space with DRoGSuRe. Our proposed approach, as a result, safeguards the private information $X_p(t), t=1,\cdots, T$ individually for each of the sensors, i.e. dedicating more degrees of freedom for each of the sensors. This is in contrast to AFDMSC. The reconstruction of the input data by the decoder, can yield the following loss function to secure the proper training of the self-expressive network, 
	\begin{equation}
		\begin{split}
			\min_{W/w_{kk}=0}\parallel W \parallel_2+ \frac{\gamma}{2} \sum_{t=1}^{T} \parallel X(t)-X_r(t)\parallel_F^2+\frac{\mu}{2} \sum_{t=1}^{T}\parallel L(t)-L(t)W\parallel_F^2,
		\end{split}
	\end{equation}
	where $\mathbf{W}$ represents the parameters of the self expressive layer, $\mathbf{X(t)}$ is the input to the encoder, $\mathbf{X}_r(t)$ denote the output of the decoder and $\mathbf{L(t)}$ denotes the output of the encoder. $\mu$ and $\gamma$ are regularization parameters. An overview for the DMSC approach is illustrated in Figure 2.
	\begin{figure}[htb]
		\centering
		\includegraphics[width=9cm,height=5cm]{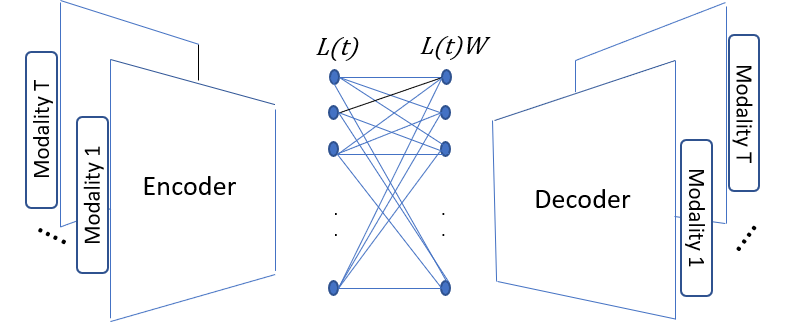}

		\caption{Deep Multimodal Subspace Clustering}.
	\end{figure}

	\section{EXPERIMENTAL RESULTS } 
	\subsection{Dataset description}
	We will evaluate our approach on two different datasets. The first dataset we will use is the Extended Yale-B dataset\cite{lee2005acquiring}. The same dataset has been used extensively in subspace clustering as in  \cite{elhamifar2013sparse,liu2012robust}. The dataset is composed of 64 frontal images of 38 individuals under different illumination conditions. In this work, we will use the augmented data used in \cite{abavisani2018deep}, where facial components such as left eye, right eye, nose and mouth have been cropped to represent four additional modalities. Images corresponding to each modality have been cropped to a size of 32$\times$32. A sample image for each modality is shown in Figure (3). The second validation dataset we use is the ARL polarimetric face dataset \cite{hu2016polarimetric}. This consists of facial images for 60 individuals in the visible domain and in four different
	polarimetric states. All the images are spatially aligned for each subject. We have also resized the images to 32$\times$32 pixels. Sample images from this dataset are shown in Figure (4).
	\begin{figure}[htb]
		\centering
		\subfloat[Face. ]{\includegraphics[width=3.5cm,height=3.1cm]{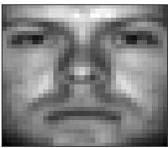}\label{fig:f1}}
		\subfloat[Left eye. ]{\includegraphics[width=3.5cm,height=3.08cm]{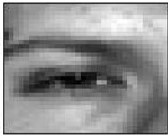}\label{fig:f2}}
		\subfloat[Right eye. ]{\includegraphics[width=3.5cm,height=3.1cm]{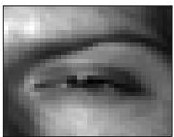}\label{fig:f3}}

		\centering
		\subfloat[Mouth. ]{\includegraphics[width=3.5cm,height=3.1cm]{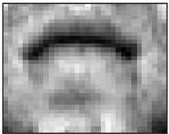}\label{fig:f4}}
		\subfloat[Nose. ]{\includegraphics[width=3.5cm,height=3.1cm]{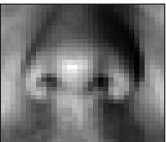}\label{fig:f5}}
		
		\caption{Sample images from the Augmented Extended Yale-B Dataset.}
	\end{figure}
	
	\begin{figure}[htb]
		\centering
		\subfloat[Visible. ]{\includegraphics[width=3.5cm,height=3.1cm]{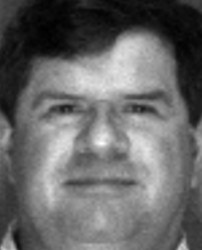}\label{fig:f6}}
		\subfloat[DoLP. ]{\includegraphics[width=3.5cm,height=3.1cm]{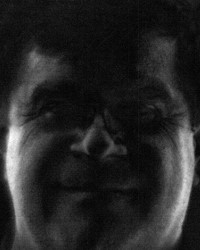}\label{fig:f7}}
		\subfloat[S0. ]{\includegraphics[width=3.5cm,height=3.1cm]{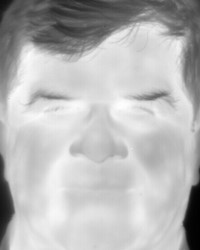}\label{fig:f8}}
		\hfill
		
		\centering
		\subfloat[S1. ]{\includegraphics[width=3.5cm,height=3.1cm]{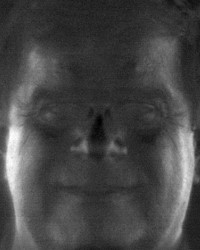}\label{fig:f9}}
		\subfloat[S2. ]{\includegraphics[width=3.5cm,height=3.1cm]{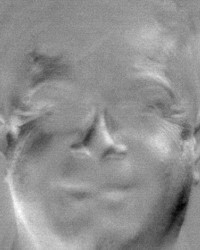}\label{fig:f10}}

		\caption{Sample images from the ARL Polarimetric Dataset.}
	\end{figure}
	
	\subsection{Network Structure}
	In the following, we will elaborate on how we construct the neural network for each dataset. Similarly to \cite{abavisani2018deep},  we implemented DRoGSuRe with Tensorflow and used the adaptive momentum based
	gradient descent method (ADAM) \cite{kingma2014adam} to minimize the
	loss function in Equation (5) with a learning rate of $10^{-3}$.
	\subsubsection{ARL Dataset}
	In case of ARL dataset, we have five data modalities and will therefore have 5 different encoders, self expressive layers and decoders. Each encoder is composed of three neural layers. The first layer consists of 5 convolutional filters of kernel size 3. The second layer has 7 filters of kernel size 1. The last layer has 15 filters with kernel size equals 1. 
	\subsubsection{EYB Dataset}
	For EYB dataset, we also have five data modalities, therefore, we have 5 different encoders, self expressive layers and decoders. Each encoder is composed of three neural layers. The first layer consists of 10 convolutional filters of kernel size 5. The second layer has 20 filters of kernel size 3. The last layer has 30 filters of kernel size 3.
	\subsection{Noiseless results}
	In the following, we compare the performance of our approach versus the DMSC approach when learning the union of subspaces structure of noise-free data. First, we divide each dataset into training and validation sets to be able to classify a newly observed dataset, using the structure learned through the current unlabeled data. The ARL expression dataset used for training consists of 2160 images per modality. The validation baseline images include 720 images total per modality. For the EYB, we randomly selected 1520 images per modality for training and 904 images for validation.The sparse solution $\mathbf{W}(t)$ corresponding to each data modality,  provides important information about the relations among data points, which may be used to split data into individual clusters residing in a common subspace. Observations from each object can be seen as data points spanning one subspace. Interpreting the subspace-based affinities based on $\mathbf{W}(t)$ as a layered set of networks, we proceed to carry out what amounts to modality fusion. The $T$ sparse matrices are added to produce one sparse matrix for both modalities, $\mathbf{W}_{Total}$, thereby improving performance.  Observations associated with one object\ individual are clustered as one subspace where the contribution of each sensor is embedded in the entries of the $\mathbf{W}_{Total}$ matrix. For clustering by $\mathbf{W}_{Total}$, we apply spectral clustering.\\
	
	After learning the structure of the data clusters, we validate our results on the validation set. We extract the principal components (eigen vectors of the covariance matrix) of each cluster in the original (training) dataset, to act as a representative subspace of its corresponding class. We subsequently project each new test point onto the subspace corresponding to each cluster, spanned by its principal components. The $l_2$ norm of the projection is then computed, and the class with the largest norm is selected to be the class of this test point. For DRoGSuRe, we use the coefficient matrix $\mathbf{W}_{total}$ in Equation (13) to cluster the test data points coming from all data modalities. We compare the clustering output labels with the ground truth for each dataset. The results for ARL and EYB datasets are depicted in Tables (1) and (2) respectively. From the results, it is clear that DRoGSuRE technique for the fused data remarkably outperforms DMSC in case of ARL dataset. However, in case of EYB dataset and in the noiseless case, DMSC performed better than DRoGSuRe. 
	\begin{table}[htbp]
		\caption{ARL dataset}
		\begin{center}
			
			\begin{tabular}{|c|c|c|c|c|}
				\hline
				
				\  & Learning & Validation    \\
				\hline 
				\ DMSC & 97.59\% & 98.33\%  \\ 
				\hline
				\ DRoGSuRe  & 100\%  &  100\% \\ 
				\hline

			\end{tabular}
			
			\label{tab1}
		\end{center}
	\end{table}  
	
	\begin{table}[htbp]
		\caption{EYB dataset}
		\begin{center}
			
			\begin{tabular}{|c|c|c|c|c|}
				\hline
				
				\  & Learning & Validation    \\
				\hline 
				\ DMSC & 98.82\% & 98.89\% \\ 
				\hline
				\ DRoGSuRe  & 98.42\% & 98.76\%  \\ 
				\hline

			\end{tabular}
			
			\label{tab2}
		\end{center}
	\end{table}  
	\subsection{Noise training with single and multiple modalities}
	In the following, we test the robustness of our approach in the case of noisy learning. We distort one modality at a time by shuffling the pixels of all images in that particular modality during the training phase. By doing so, we are perturbing the structure of the sparse coefficient matrix associated with that modality, thus impacting the overall W matrix for both DRoGSuRe and DMSC. Testing with clean data, i.e. no distortion, demonstrates the impact of perturbing the training and hence performing an inadequate training, e.g., insufficient data or non-convergence. This can also be considered as augmenting the training data with new information or a new view for one modality which might not necessarily contained in the testing or the validation data. Moreover, we repeat the same experiment with the distortion of two modalities before learning the sparse coefficient matrices for both DMSC and DRoGSuRe. The results for the ARL dataset are depicted in Table (3) and (4), while results for the EYB dataset are shown in Table (5) and (6). For ARL dataset, we refer to Visible, S0, S1, S2 and DoLP as Mod 0, 1, 2, 3 and 4 respectively. For EYB Dataset, we refer to Face, left eye, nose, mouth and right eye as mod 0, 1, 2, 3, and 4.  We refer to each modality as Mod, where L denotes learning and V denotes validation results. From the results, it is clear that DRoGSuRe is showing a significant improvement in the clustering accuracy as compared to DMSC for both learning and validation set. The reason for that, is again, due to the fact that perturbing one or two modalities would have less impact on the overall performance for DRoGSuRe in comparison to DMSC.
	\begin{table}[htbp]
		\caption{ARL data :Distorting One Modality}
		\begin{center}
			
			\begin{tabular}{|c|c|c|c|c|}
				\hline
				
				& DMSC L & DMSC V & DRoGSuRe L & DRoGSuRe V   \\
				
				\hline 
				
				Mod 0 & 87.17\%  &86.67 \% & 95.37\% & 95\%  \\ 
				\hline
				Mod 1 & 91.67\%  & 90\% & 98.29\%  & 98.33\%  \\ 
				\hline 
				Mod 2 & 92.77\%  &  92.78\% & 99.17\%  & 99.44\%  \\ 
				\hline 
				Mod 3 & 90.55\%  & 90.57\% & 99.31\%  & 99.44\%  \\ 
				\hline 
				Mod 4 & 92.78\%  & 91.11\% & 96.44\%  & 96.67\%  \\ 
				\hline 
				
			\end{tabular}
			
			\label{tab3}
		\end{center}
	\end{table}
	
	\begin{table}[htbp]
		\caption{ARL data :Distorting Two Modalities}
		\begin{center}
			
			\begin{tabular}{|c|c|c|c|c|}
				\hline
				
				& DMSC L & DMSC V & DRoGSuRe L & DRoGSuRe V   \\
				
				\hline 
				
				Mod 0 \& 1 & 82.22\%  & 82.78\% & 92.27\% & 94.58\%  \\ 
				\hline
				Mod 1 \& 2 & 91.11\%  & 91.11\% & 97.22\%  & 97.36\%  \\ 
				\hline 
				Mod 0 \& 3 & 85.51\%  & 82.56\% & 93.01\%  & 95.42\%  \\ 
				\hline 
				Mod 1 \& 4 & 91.67\%  & 89.44\% & 97.22\%  & 97.36\%  \\ 
				\hline 
				Mod 2 \& 3 & 90\%  & 89.72\% & 97.69\%  & 97.78\%  \\ 
				\hline 
				
			\end{tabular}
			
			\label{tab4}
		\end{center}
	\end{table}
	
	\begin{table}[htbp]
		\caption{EYB data :Distorting One Modality}
		\begin{center}
			
			\begin{tabular}{|c|c|c|c|c|}
				\hline
				
				& DMSC L & DMSC V & DRoGSuRe L & DRoGSuRe V   \\
				
				\hline 
				
				Mod 0 & 87.96\%  & 88.5\% & 93.29\% & 94.69\%  \\ 
				\hline
				Mod 1 & 91.84\%  & 91.15\% & 95.79\%  & 97.46\%  \\ 
				\hline 
				Mod 2 & 89.01\%  & 88.72\% & 98.03\%  & 97.57\%  \\ 
				\hline 
				Mod 3 & 92.69\%  & 91.81\% & 95.59\%  & 96.68\%  \\ 
				\hline 
				Mod 4 & 91.45\%  & 91.59\% & 97.17\%  & 97.35\%  \\ 
				\hline 
				
			\end{tabular}
			
			\label{tab5}
		\end{center}
	\end{table}

	\begin{table}[htbp]
		\caption{EYB data :Distorting Two Modalities}
		\begin{center}
			
			\begin{tabular}{|c|c|c|c|c|}
				\hline
				
				& DMSC L & DMSC V & DRoGSuRe L & DRoGSuRe V   \\
				
				\hline 
				
				Mod 0 \& 2 & 86.64\%  & 85.18\% & 96.84\% & 96.13\%  \\ 
				\hline
				Mod 0 \& 4 & 87.83\%  & 89.16\% & 94.54\%  & 95.8\%  \\ 
				\hline 
				Mod 1 \& 4 & 86.38\%  & 86.06\% & 94.21\%  & 95.8\%  \\ 
				\hline 
				Mod 2 \& 3 & 88.22\%  & 84.96\% & 91.58\%  & 93.92\%  \\ 
				\hline 
				Mod 3 \& 4 & 88.03\%  & 86.28\% & 94.08\%  & 95.35\%  \\ 
				\hline 
				
			\end{tabular}
			
			\label{tab6}
		\end{center}
	\end{table}
	
	\subsection{Testing with limited noisy testing data}
	In the following, we study the effect of using noiseless data for training while validating with noisy and missing data. We add Gaussian noise to one data modality in the validation set, and vary the SNR by varying the noise variance. We subsequently assume that we only have one modality available at testing. Then, we keep increasing the number of available noiseless data modalities beside the noisy modality. We average the results considering all different combinations of data modalities for ARL and EYB datasets. The results are depicted in Figure (5) and (6) respectively. For the ARL dataset, we note the increasing gap between DMSC and DRoGSuRe as we augment the sensing capacity with noise-free modalities. On the other hand, for the EYB dataset and at lower SNR, the performance of DRoGSuRe is slightly worse than DMSC which might be explained by the results in Table II; as the training accuracy for DMSC is slightly better than DRoGSuRe in the case of clean training. However, at higher SNR, the performance of the two approaches is very close. 
	
	\begin{figure}
		\centering
		\includegraphics[width=8cm,height=5cm]{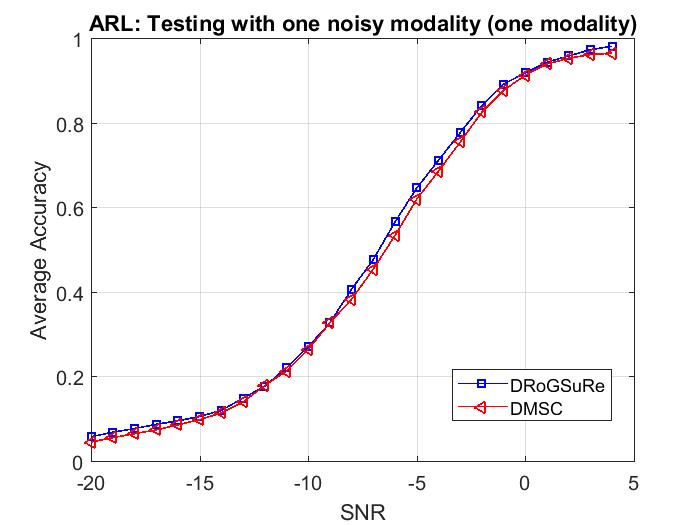}
		\includegraphics[width=8cm,height=5cm]{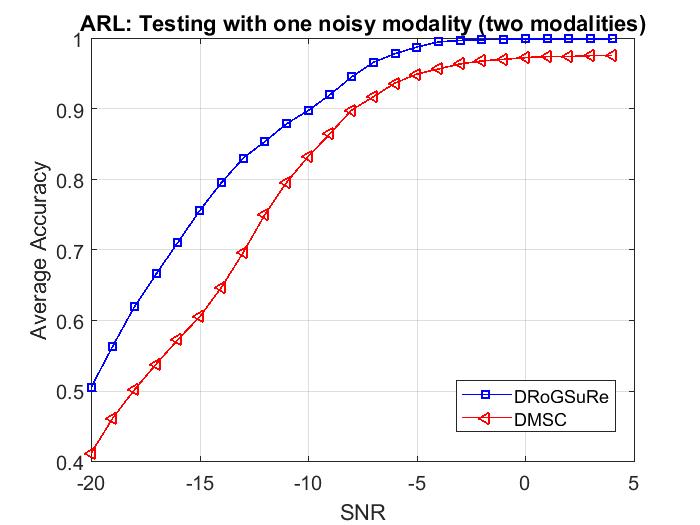}
		
		\centering
		\includegraphics[width=8cm,height=5cm]{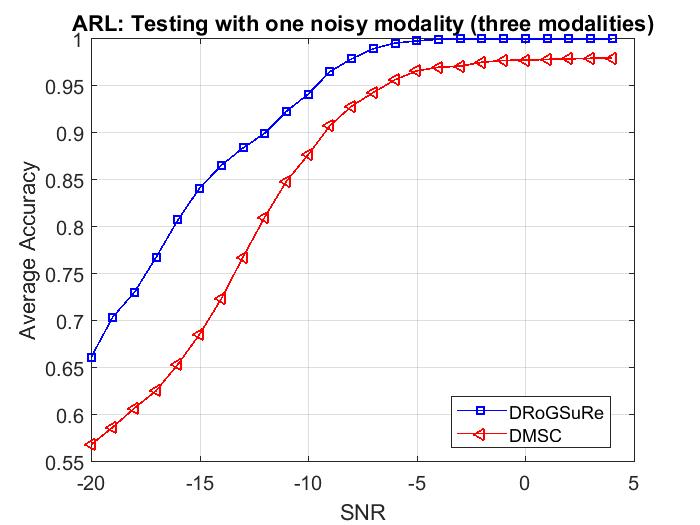}
		\includegraphics[width=8cm,height=5cm]{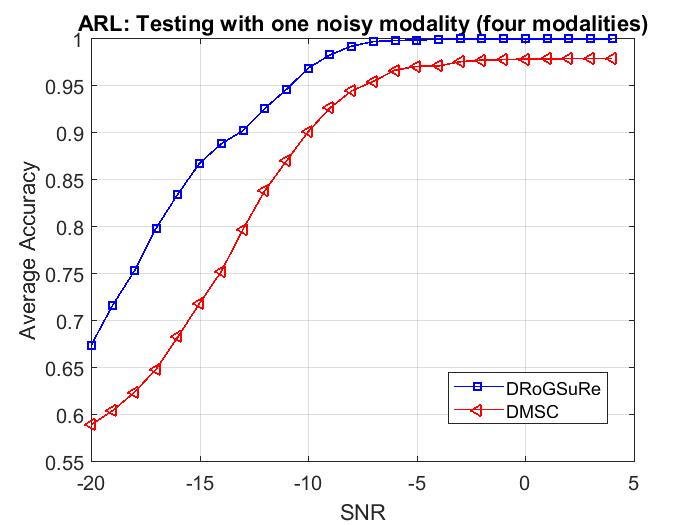}
		
		\caption{ARL Noiseless Training and validating on limited noisy data.}
	\end{figure}
	
	\begin{figure}
		\centering
		\includegraphics[width=8cm,height=5cm]{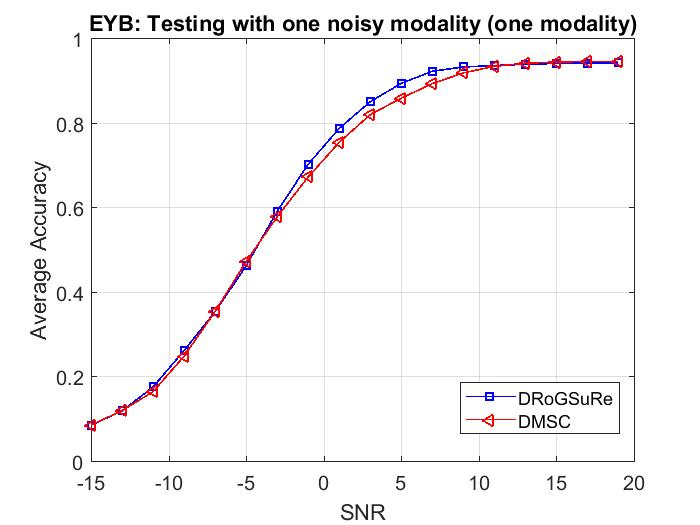}
		\includegraphics[width=8cm,height=5cm]{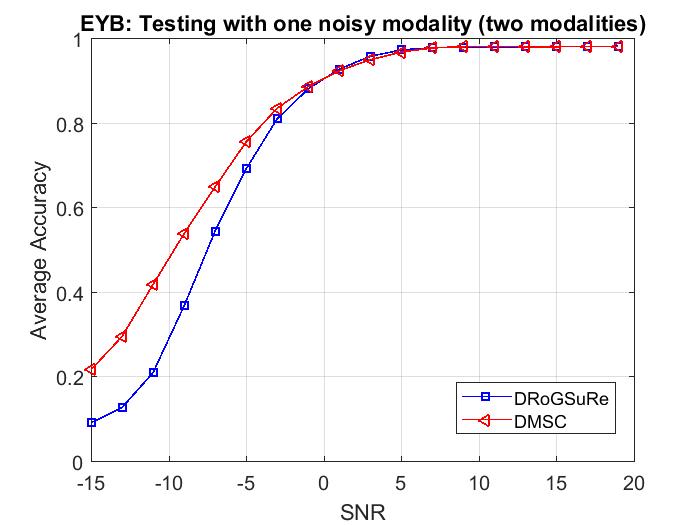} 
		
		\centering
		\includegraphics[width=8cm,height=5cm]{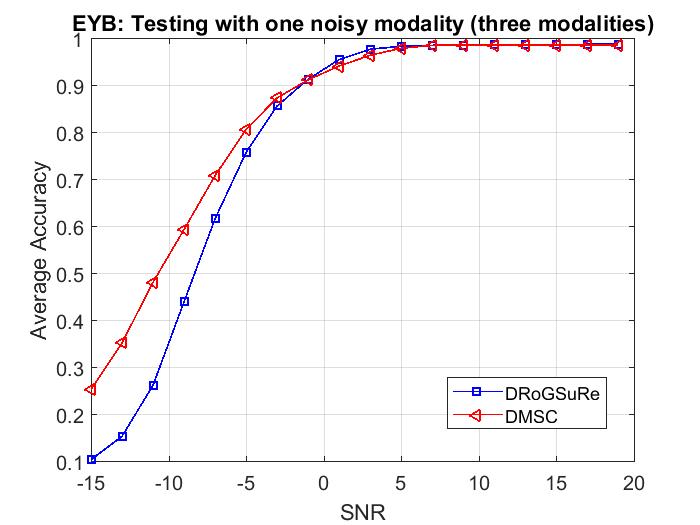}
		\includegraphics[width=8cm,height=5cm]{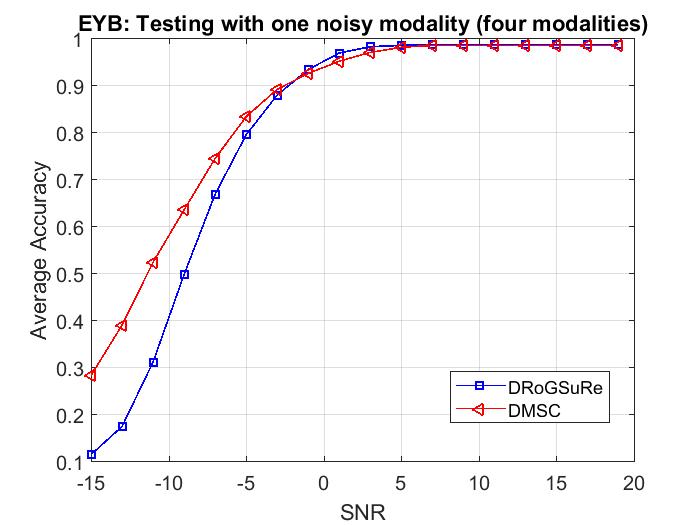}
		
		\caption{EYB Noiseless Training and validating on limited noisy data.}
	\end{figure}
	
	\subsection{Missing modalities during testing}
	In the following, we evaluate the performance of DRoGSuRe and DMSC in case of missing data modalities during testing. It is not uncommon to have one or more sensors that  might be silent during testing, thus justifying this experiment for further assessment. We try different combinations of available modalities during testing, and we average the clustering accuracy for each trial. Results are depicted in Figures (7) and (8) for ARL and EYB data respectively. Again, we notice a significant improvement for DRoGSuRe over DMSC for ARL Dataset. For EYB dataset, there is a slight improvement for DRoGSuRe over DMSC, however, the performance of DRoGSuRe is gracefully degrading while more modalities become unavailable during testing, which emphasizes the reliability and robustness of our proposed approach. 
	\begin{figure}[htb]
		\centering
		{\includegraphics[width=10cm,height=7cm]{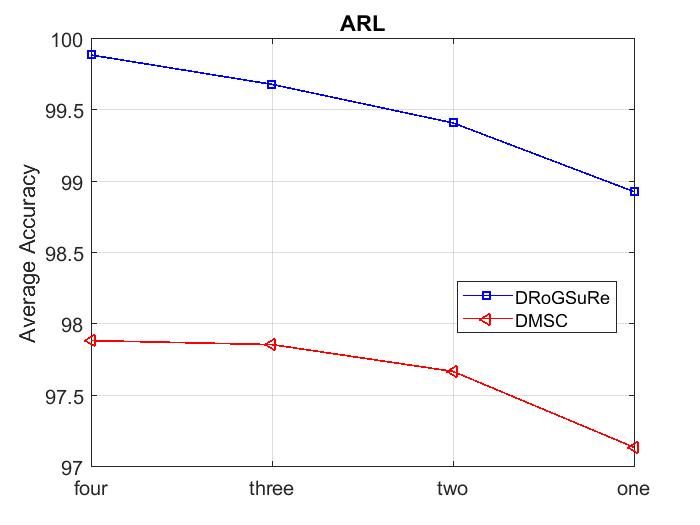}\label{fig:f29}}
		\caption{Missing modalities during testing for ARL dataset.}
	\end{figure}
	\begin{figure}[htb]
		\centering
		{\includegraphics[width=10cm,height=7cm]{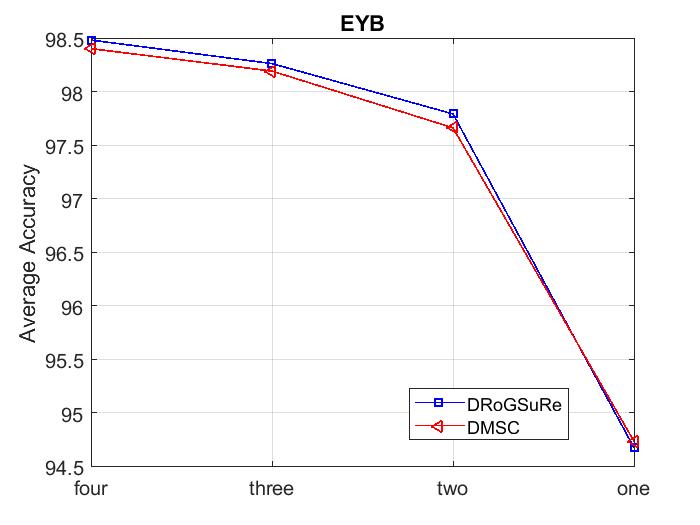}\label{fig:f30}}

		\caption{Missing modalities during testing for EYB dataset.}
	\end{figure}
	
	
	\section{Feature Concatenation}
	Here we propose a rationale along with an alternative solution for enhancing the performance for EYB multi-modal data. Due to the specific structure of the EYB multi-modal data, the concatenation of the features corresponding to each modality is a reasonable alternative. By doing so, we are adjoining together the features representing each part of the face. Since the four modalities correspond to non-overlapping partitions of the face, the feature set corresponding to each partition will solely provide complementing information. A similar idea is proposed in \cite{abavisani2018deep} and is referred to as Late concatenation, where the multi-modal data is integrated in the last stage of the encoder. Their resulting decoder structure remains the same for either affinity fusion or late concatenation. This entails deconcatenating the multi-modal data prior to decoding it. Our proposed approach on the other hand, results in a self-expressive layer being driven by the concatenated features from the $M$ encoder branches. Afterwards, we feed the self-expressive layer output to each branch of the decoder. The concatenated information results in a more efficient code for the data, thereby resulting in an overall parsimonious with a sparse structure of the decoder, results in a decoder composed of three neural layers. The first layer consists of 150 filters of kernel size 3. The second layer consists of 20 layers of kernel size 3. The third layer consists of 10 layers of kernel size 5. Our approach is illustrated in Figure (9). We optimize the weights of the auto-encoder as follows,
	\begin{equation}
		\begin{split}
			\min\limits_{\mathbf{W} \mid w_{kk}=0} \rho \parallel \mathbf{W} \parallel_1+
			\frac{\gamma}{2} \sum_{t=1}^{T} \parallel X(t)-X_r(t) \parallel_F^2 +\frac{\mu}{2} \parallel N-NW \parallel_F^2.
		\end{split}  
	\end{equation} 
	where $N=[L(1)||L(2)||L(3)||L(4)||L(5)]$
	\begin{figure}[htb]
		\centering
		\includegraphics[width=13cm,height=5cm]{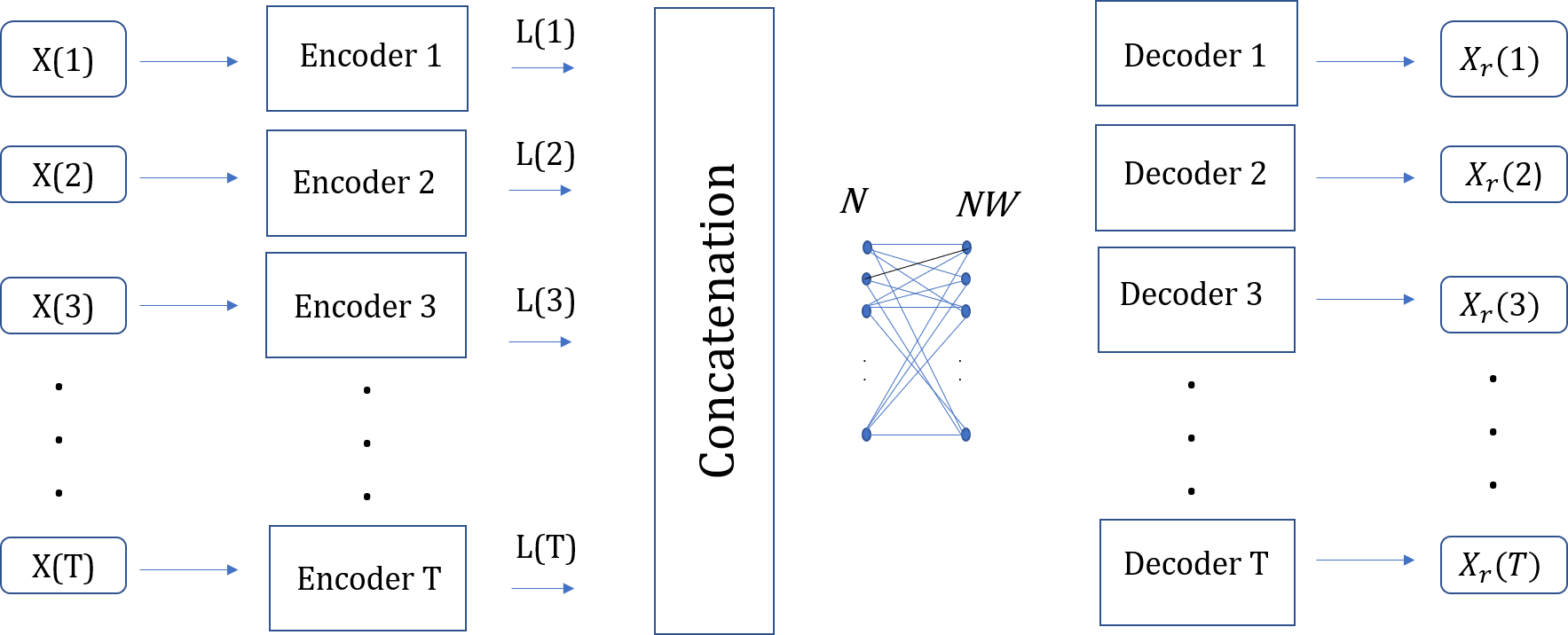}

		\caption{CNN Concatenation Network}.
	\end{figure}
	We compared the performance of our proposed approach against the late concatenation approach in \cite{abavisani2018deep} and the results are depicted in Table 7 for the EYB dataset.
	\begin{table}[htbp]
		\caption{EYB dataset}
		\begin{center}
			
			\begin{tabular}{|c|c|c|c|c|}
				\hline
				
				\  & Learning & Validation    \\
				\hline 
				\ DMSC Late Concatenation & 95.66\% & 94.7\% \\ 
				\hline
				\ CNN Concatenation Network   & 99.28\% & 99.3\%   \\ 
				\hline

			\end{tabular}
			
			\label{tab7}
		\end{center}
	\end{table}  
	From the previous table, we can conclude that concatenating the features from the encoder and feeding the concatenated information to each decoder branch achieves a better performance for this type of multi-modal data structure. The reason behind this enhancement is the combination of  efficient extraction of the basic features from the whole face and finer features from each part of the face. Promoting  more efficiency as noted, this concatenation  may also be intuitively viewed as adequate mosaicking, in which different patterns complement each other. In the following, we will show how our proposed approach performs in two cases; missing and noisy test data. The results of the new proposed approach, which we refer to as CNN concatenation network, is compared to the state-of-the-art DMSC network \cite{abavisani2018deep}. We start by training the auto-encoder network using 75\% of the data and then we test on the rest of the data. In Figure 10, we show how the performance degrades by decreasing the number of available modalities at testing from five to one.  From the results, it is clear how the CNN concatenation network outperforms the DMSC network. Additionally, we repeated the same experiment we performed in subsection 4.5. We train the network with noiseless data and then add Gaussian noise to one data modality at the testing. Additionally, we vary the number of available modalities at testing from one to four. The results are depicted in Figure 11.  From the results, it is clear how the concatenated CNN network is more robust to noise than DMSC.
	\begin{figure}[htb]
		\centering
		\includegraphics[width=8.5cm,height=6cm]{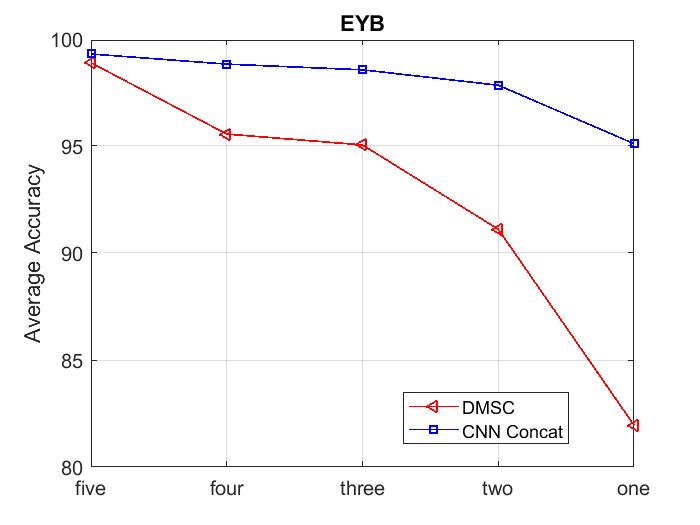}

		\caption{Missing modalities during testing}.
	\end{figure} 
	
	\begin{figure}
		\centering
		\includegraphics[width=8cm,height=5cm]{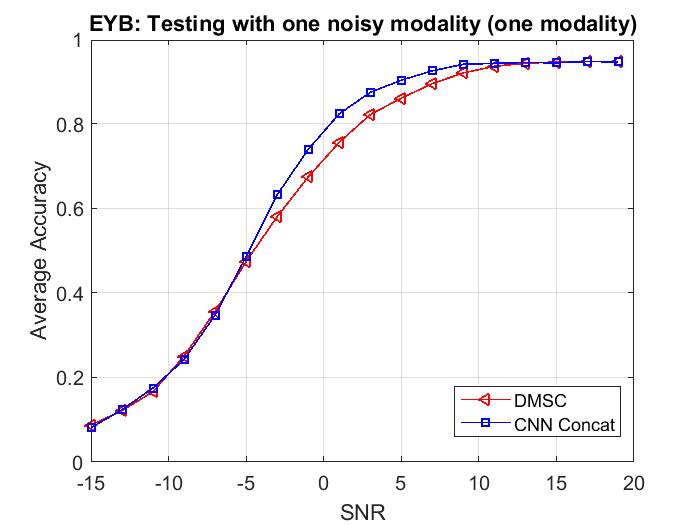}
		\includegraphics[width=8cm,height=5cm]{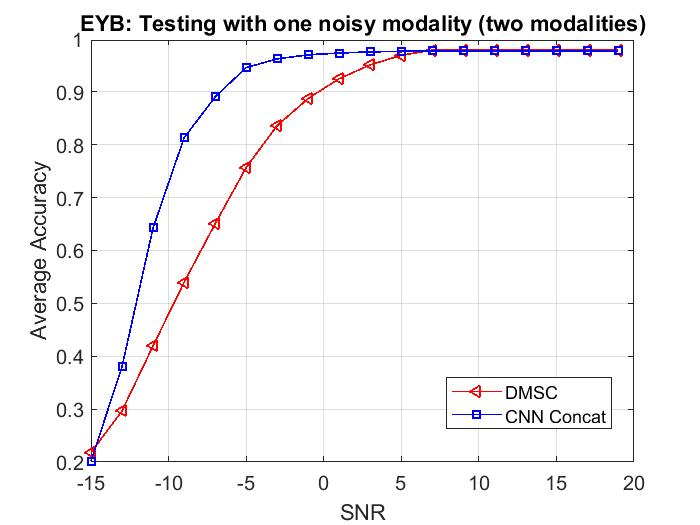} 
		\centering
		\includegraphics[width=8cm,height=5cm]{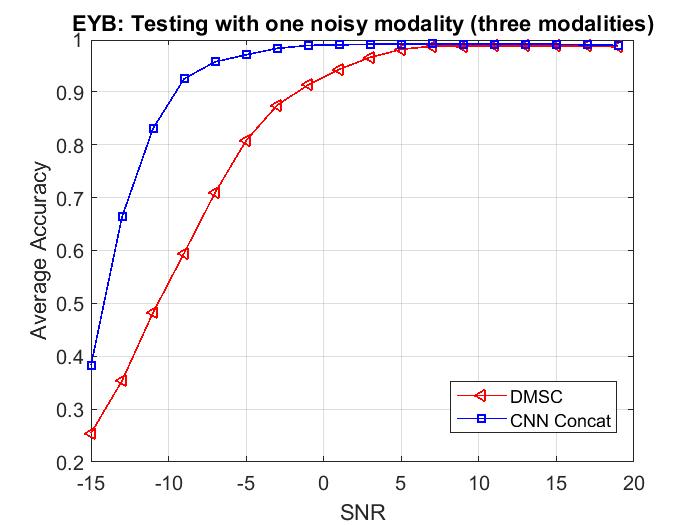}
		\includegraphics[width=8cm,height=5cm]{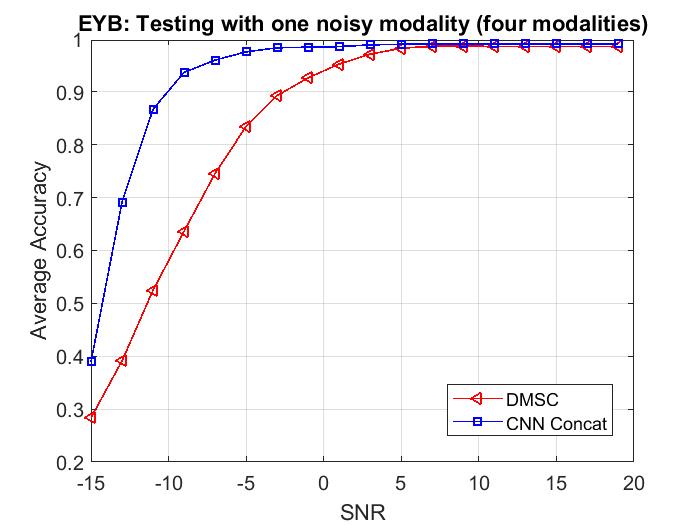}
		
		\caption{EYB Noiseless Training and validating on limited noisy data.}
	\end{figure}
	In addition, we have utilized the Concatenation network to perform object clustering on the ARL data. We compare the clustering performance of the concatenation network with both DMSC and DRoGSuRe. The results are depicted in Table 8. From the results, we conclude that DRoGSuRe still outperforms the other approaches for the ARL dataset. Although the number of parameters involved in training the DRoGSuRe network is higher than other approaches, since there are multiple self expressive layers, however, DRoGSuRe is more robust to noise and limited data availability during testing. 
	\begin{table}[htbp]
		\caption{ARL dataset}
		\begin{center}
			
			\begin{tabular}{|c|c|c|c|c|}
				\hline
				
				\  & Learning & Validation    \\
				\hline 
				\ DMSC & 97.59\% & 98.33\%  \\ 
				\hline
				\ DRoGSuRe  & 100\%  &  100\% \\ 
				\hline 
				\ CNN Concat  & 99.44\%  & 99.17\%   \\ 
				\hline

			\end{tabular}
			
			\label{tab11}
		\end{center}
	\end{table}  
	
	\subsection{Deeper Network}
	In this work, we addressed the multi-modal data fusion by building on Deep Subspace Clustering, and we compare our approach to deep multimodal subspace\footnote{This work was pointed out to us by a reviewer in the course of the review, and we have added an experiment for a further and fair evaluation.} clustering network \cite{abavisani2018deep} which in some sense dictated our following the same network structure in their paper so that we can have a fair comparison with their approach. In this section, we will consider a deeper alternative for the concatenation network and  evaluate its performance. In Tables 7 and 8, we showed the results for the 3-layered concatenation network using ARL and EYB, respectively. Both the encoder and the decoder had three layers. We considered using 4, 5, and 6 layers in the auto-encoder for both the ARL and EYB to test deeper networks. The results are listed in Table 9. From the results, it is clear that adding more layers to the network does not improve the classification accuracy, which we believe is due to the dimensionality of the images in both datasets. All images are 32 $\times$ 32, and might not need additional layers to capture the image details.   
	
	\begin{table}[htbp]
		\caption{Deep Network}
		\begin{center}
			
			\begin{tabular}{|c|c|c|c|c|c|c|c|c|c|c|}
				\hline
				
				\  & \multicolumn{3}{c|}{4 layers} & \multicolumn{3}{c|}{5 layers} & \multicolumn{3}{c|}{6 layers}      \\
				\hline
				
				\  & ACC & ARI & NMI & ACC & ARI & NMI & ACC & ARI & NMI     \\
				\hline 
				\ ARL & 99.4\% & 99.67\% & 98.94\% & 99.31\% & 99.6\% & 98.76\% & 99.26\% & 99.55\% & 98.66\%  \\ 
				\hline
				\ EYB  & 99.14\%  & 98.9\% & 98.23\% & 97.43\% & 97.13\% & 94.77\% & 63.88\% & 58.31\% & 68.83\%  \\ 
				\hline

			\end{tabular}
			
			\label{tab12}
		\end{center}
	\end{table}   
	\section{Conclusion}
	
	In this paper, we proposed a deep multi-modal approach to fuse data through recovering the underlying subspaces of data observations from data corrupted by noise to scale to complex data scenarios. DRoGSuRe provides a natural way to fuse multi-modal data by employing the self representation matrix as an embedding for each data modality. Experimental results show a significant improvement for DRoGSuRe over DMSC under different types of potential limitations and provides robustness with limited sensing modalities. We also proposed the concatenated CNN network model, which can work better for different multi-modal data structures.

	\appendix
	\section{Parameter Perturbation Analysis}
	To theoretically compare our proposed variational scaling fusion approach DRoGSuRe to DMSC, we proceed by way of a first order perturbation analysis on the parameter set $W^i$of respectively either technique $i=$1,2. This will, in turn impact the associated affinity matrix $A^i$, which as we will later elaborate directly impacts the subspace clustering procedure which is central to the inference following the fusion procedure. 
	
	Adopting the original formulation for the first persistently differential scaling approach, namely that T modalities are jointly exploited, results in, $X^1(t)=[x_1^1(t) x_2^1(t) ... x_n^1(t)]$, where $x_k^1(t) \in  \mathbb{R}^m, t=1,2,...,T $ represents the $k^{th}$ observation.  The second approach only effectively uses only one subspace structure of the fused modalities$X^2(t)=[x_1^2 x_2^2 ... x_n^2]$.
	A first order perturbation on the data may be due to noise or to a degradation of a given sensor, and results in a perturbation of the UoS parameters,
	\begin{equation}
		\tilde{W}^i_1=W^i_1+\delta^i
	\end{equation} 
	For the first method, each modality will have an associated subspace cluster parameter set $\{W_t^1\}_{t=1,...,T}$, with  $W_t^1 \in \mathbb{R}^{n \times n}$. The overall parameter set for DRoGSuRe can then be written as,
	\begin{equation}
		\tilde{W}^1=\tilde{W}^1_1+W^1_2+...+W_m^1
	\end{equation}
	Where the unperturbed overall sparse coefficient matrix is written as follows, $ W_{tot}^1=W_1^1+W_2^1+...+W_m^1$. A similar development follows for method 2, with the difference that the contributing modalities are fused a priori.  
	\subsubsection*{Proof}
	We first write the affinity matrix associated with DRoGSuRE as, 
	\begin{equation}
		\tilde{A}^1=\tilde{W}^1_{tot}+(\tilde{W}^1_{tot})^T
	\end{equation}
	\begin{equation}
		\tilde{A}^1=\tilde{W}^1_1+W^1_2+...+W_m^1+(\tilde{W}^1_1+W^1_2+...+W_m^1)^T
	\end{equation}
	Where the superscript $T$ denotes transpose. This is equivalent to,
	\begin{equation}
		\tilde{A}^1=\tilde{A}^1_1+\sum_{i=2}^{T} A_i^1,
	\end{equation}
	Where $0\leq \tilde{A}_1^1(i,j) \leq 1+	\delta^1$. The unperturbed collective affinity matrix $A^1$  can be similarly written $A^1=\sum_{i=1}^{T} A_i^1$   with the unity constraint on each entry of all  matrices. We may also write the magnitude of the difference as, 
	\begin{equation}
		\lvert A^1-\tilde{A}^1\rvert=\delta^1+(\delta^1)^T
	\end{equation}
	Letting $\Delta=\delta^1+(\delta^1)^T \in \mathbb{R}^{n \times n} $, i.e., $\Delta \in \mathbb{R}^{n \times n}$ and assuming $\epsilon=\max_{i,j}[\Delta]_{i,j}$, we can write,
	\begin{equation}
		\parallel  A-\tilde{A}\parallel_F \leq n\epsilon
	\end{equation}
	Given the $\Delta$ matrix individual entry bounds , we conclude that,
	\begin{equation}
		0\leq \epsilon \leq \frac{1}{t}
	\end{equation}
	Since DMSC assumes having one sparse coefficient matrix W for all data modalities , which is equivalent to only one subspace structure of the fused modalities $X^2(t)=[x^2_1 ... x_n^2]$. Therefore, the UoS parameters will be perturbed by $\delta^2$ as follows,
	\begin{equation}
		\tilde{W}^2=W^2+\delta^2	
	\end{equation}
	The affinity matrix associated with DMSC can be written as follows, $\tilde{A}^2=\tilde{W}^2+(\tilde{W}^2)^T$, which is equivalent to,
	\begin{equation}
		\tilde{A}^2=W^2+\delta^2+(W^2)^T+(\delta^2)^T
	\end{equation}
	Similarly, the unperturbed affinity matrix will be as follows,
	
	\begin{equation}
		A^2=W^2+(W^2)^T
	\end{equation}
	From Eqns. 28 and 29, the magnitude of the difference can be written as follows, 
	\begin{equation}
		\lvert A^2-\tilde{A}^2 \rvert=\delta^2+(\delta^2)^T
	\end{equation}
	Letting $\gamma=\delta^2+(\delta^2)^T \in \mathbb{R}^{n \times n}$, i.e., $\lvert A^2-\tilde{A}^2 \rvert=\gamma $, and assuming  $ \psi =\max_{i,j}[\gamma]_{ij}$, we can write $\parallel A^2-\tilde{A}^2\parallel_F \leq n\psi$ .          
	Given the $ \gamma $ matrix individual entry bounds, we conclude $0 \leq \psi \leq 1$.          
	If we only perturb one modality, knowing that  $0 \leq A(i,j) \leq 1$, therefore the error could lie between $0 \leq \psi \leq 1$, which entails either creating a fake relation between two data points or erasing an existing relation. $\epsilon$ and $\psi$ are random variables that do not have to follow a specific distribution, however, in any case $E(\epsilon^2 )\ll E(\psi^2) $and therefore $ SNR_{DRoGSuRe} \gg SNR_{DMSC}$
	
	In light of the above two bounds, and the results of \cite{hunter2010performance}, where it is shown that the spectral clustering dependent on the respective projection operators $P_{W^1}$ and $\tilde{P}_{\tilde{W}^1}$ onto the vector subspaces spanned by the principal eigenvectors of $W^1_{tot}$ and $\tilde{W}^1_{tot}$ of may be written as,
	\begin{equation}
		\parallel P_{W^1}-\tilde{P}_{\tilde{W}^1}\parallel_F \leq \frac{\sqrt{2}}{\alpha^1} \parallel A^1-\tilde{A}^1\parallel_F,
	\end{equation}
	where $\alpha^1$ is the spectral gap between the $k^{th}$ and $k+1^{st}$ eigen value of $A^1$, $\lvert \lambda^1_k-\lambda^1_{k+1} \rvert$. Similarly, for DMSC, the bound on the projection operators is,
	\begin{equation}
		\parallel P_{W^2}-\tilde{P}_{\tilde{W}^2}\parallel_F \leq \frac{\sqrt{2}}{\alpha^2} \parallel A^2-\tilde{A}^2\parallel_F,
	\end{equation}
	where $\alpha^2=\lvert \lambda^1_k-\lambda^1_{k+1} \rvert$. Since $W_1^1,W_2^1,..., W_T^1$  happen to commute and if they happen to be diagonalizable, therefore, they share the same eigenvectors. As a result, the eigenvectors of $W_1^1+ W_2^1+...+W_m^1 $ are also the same and the corresponding eigenvalue that is the sum of the corresponding eigenvalues of $W_1^1,  W_2^1,..$ and $W_T^1$. As a result, we conclude that $\lambda_k^2$ Therefore, $\lambda_k^1 \gg \lambda_k^2$. From all the above, we can conclude that smaller error yielding to better clustering, hence preserving the performance, yields the improvement by the T-factor noted in the proposition and shown in the two perturbation developments.
	\section*{Acknowledgment}

	This work was in part supported by DOE-National Nuclear Security Administration through CNEC-NCSU under Award DE-NA0002576. The first author was also in part supported by DTRA. 
	\printcredits
	
	\bibliographystyle{cas-model2-names}
	
	\bibliography{deep_paper}

\begin{thebibliography}{40}
\expandafter\ifx\csname natexlab\endcsname\relax\def\natexlab#1{#1}\fi
\providecommand{\url}[1]{\texttt{#1}}
\providecommand{\href}[2]{#2}
\providecommand{\path}[1]{#1}
\providecommand{\DOIprefix}{doi:}
\providecommand{\ArXivprefix}{arXiv:}
\providecommand{\URLprefix}{URL: }
\providecommand{\Pubmedprefix}{pmid:}
\providecommand{\doi}[1]{\href{http://dx.doi.org/#1}{\path{#1}}}
\providecommand{\Pubmed}[1]{\href{pmid:#1}{\path{#1}}}
\providecommand{\bibinfo}[2]{#2}
\ifx\xfnm\relax \def\xfnm[#1]{\unskip,\space#1}\fi
\bibitem[{Abavisani and Patel(2018)}]{abavisani2018deep}
\bibinfo{author}{Abavisani, M.}, \bibinfo{author}{Patel, V.M.},
  \bibinfo{year}{2018}.
\newblock \bibinfo{title}{Deep multimodal subspace clustering networks}.
\newblock \bibinfo{journal}{IEEE Journal of Selected Topics in Signal
  Processing} \bibinfo{volume}{12}, \bibinfo{pages}{1601--1614}.
\bibitem[{Bian and Krim(2015)}]{bian2015bi}
\bibinfo{author}{Bian, X.}, \bibinfo{author}{Krim, H.}, \bibinfo{year}{2015}.
\newblock \bibinfo{title}{Bi-sparsity pursuit for robust subspace recovery},
  in: \bibinfo{booktitle}{2015 IEEE International Conference on Image
  Processing (ICIP)}, \bibinfo{organization}{IEEE}. pp.
  \bibinfo{pages}{3535--3539}.
\bibitem[{Bian et~al.(2018)Bian, Panahi and Krim}]{bian2018bi}
\bibinfo{author}{Bian, X.}, \bibinfo{author}{Panahi, A.},
  \bibinfo{author}{Krim, H.}, \bibinfo{year}{2018}.
\newblock \bibinfo{title}{Bi-sparsity pursuit: A paradigm for robust subspace
  recovery}.
\newblock \bibinfo{journal}{Signal Processing} \bibinfo{volume}{152},
  \bibinfo{pages}{148--159}.
\bibitem[{Chen and Hero(2017)}]{chen2017multilayer}
\bibinfo{author}{Chen, P.Y.}, \bibinfo{author}{Hero, A.O.},
  \bibinfo{year}{2017}.
\newblock \bibinfo{title}{Multilayer spectral graph clustering via convex layer
  aggregation: Theory and algorithms}.
\newblock \bibinfo{journal}{IEEE Transactions on Signal and Information
  Processing over Networks} \bibinfo{volume}{3}, \bibinfo{pages}{553--567}.
\bibitem[{Dong et~al.(2013)Dong, Frossard, Vandergheynst and
  Nefedov}]{dong2013clustering}
\bibinfo{author}{Dong, X.}, \bibinfo{author}{Frossard, P.},
  \bibinfo{author}{Vandergheynst, P.}, \bibinfo{author}{Nefedov, N.},
  \bibinfo{year}{2013}.
\newblock \bibinfo{title}{Clustering on multi-layer graphs via subspace
  analysis on grassmann manifolds}.
\newblock \bibinfo{journal}{IEEE Transactions on signal processing}
  \bibinfo{volume}{62}, \bibinfo{pages}{905--918}.
\bibitem[{Elhamifar and Vidal(2013)}]{elhamifar2013sparse}
\bibinfo{author}{Elhamifar, E.}, \bibinfo{author}{Vidal, R.},
  \bibinfo{year}{2013}.
\newblock \bibinfo{title}{Sparse subspace clustering: Algorithm, theory, and
  applications}.
\newblock \bibinfo{journal}{IEEE transactions on pattern analysis and machine
  intelligence} \bibinfo{volume}{35}, \bibinfo{pages}{2765--2781}.
\bibitem[{Favaro et~al.(2011)Favaro, Vidal and Ravichandran}]{favaro2011closed}
\bibinfo{author}{Favaro, P.}, \bibinfo{author}{Vidal, R.},
  \bibinfo{author}{Ravichandran, A.}, \bibinfo{year}{2011}.
\newblock \bibinfo{title}{A closed form solution to robust subspace estimation
  and clustering}, in: \bibinfo{booktitle}{CVPR 2011},
  \bibinfo{organization}{IEEE}. pp. \bibinfo{pages}{1801--1807}.
\bibitem[{Gabriel et~al.(2010)Gabriel, Spiliopoulou and
  Nanopoulos}]{gabriel2010eigenvector}
\bibinfo{author}{Gabriel, H.H.}, \bibinfo{author}{Spiliopoulou, M.},
  \bibinfo{author}{Nanopoulos, A.}, \bibinfo{year}{2010}.
\newblock \bibinfo{title}{Eigenvector-based clustering using aggregated
  similarity matrices}, in: \bibinfo{booktitle}{Proceedings of the 2010 ACM
  Symposium on Applied Computing}, pp. \bibinfo{pages}{1083--1087}.
\bibitem[{Ghanem et~al.(2020)Ghanem, Panahi, Krim and
  Kerekes}]{ghanem2020robust}
\bibinfo{author}{Ghanem, S.}, \bibinfo{author}{Panahi, A.},
  \bibinfo{author}{Krim, H.}, \bibinfo{author}{Kerekes, R.A.},
  \bibinfo{year}{2020}.
\newblock \bibinfo{title}{Robust group subspace recovery: A new approach for
  multi-modality data fusion}.
\newblock \bibinfo{journal}{IEEE Sensors Journal} .
\bibitem[{Ghanem et~al.(2018)Ghanem, Panahi, Krim, Kerekes and
  Mattingly}]{ghanem2018information}
\bibinfo{author}{Ghanem, S.}, \bibinfo{author}{Panahi, A.},
  \bibinfo{author}{Krim, H.}, \bibinfo{author}{Kerekes, R.A.},
  \bibinfo{author}{Mattingly, J.}, \bibinfo{year}{2018}.
\newblock \bibinfo{title}{Information subspace-based fusion for vehicle
  classification}, in: \bibinfo{booktitle}{2018 26th European Signal Processing
  Conference (EUSIPCO)}, \bibinfo{organization}{IEEE}. pp.
  \bibinfo{pages}{1612--1616}.
\bibitem[{Ghanem et~al.(2021)Ghanem, Roheda and Krim}]{ghanem2021latent}
\bibinfo{author}{Ghanem, S.}, \bibinfo{author}{Roheda, S.},
  \bibinfo{author}{Krim, H.}, \bibinfo{year}{2021}.
\newblock \bibinfo{title}{Latent code-based fusion: A volterra neural network
  approach}.
\newblock \bibinfo{journal}{arXiv preprint arXiv:2104.04829} .
\bibitem[{Hellwich and Wiedemann(2000)}]{hellwich2000object}
\bibinfo{author}{Hellwich, O.}, \bibinfo{author}{Wiedemann, C.},
  \bibinfo{year}{2000}.
\newblock \bibinfo{title}{Object extraction from high-resolution multisensor
  image data}, in: \bibinfo{booktitle}{Third International Conference Fusion of
  Earth Data, Sophia Antipolis}.
\bibitem[{Ho et~al.(2003)Ho, Yang, Lim, Lee and Kriegman}]{ho2003clustering}
\bibinfo{author}{Ho, J.}, \bibinfo{author}{Yang, M.H.}, \bibinfo{author}{Lim,
  J.}, \bibinfo{author}{Lee, K.C.}, \bibinfo{author}{Kriegman, D.},
  \bibinfo{year}{2003}.
\newblock \bibinfo{title}{Clustering appearances of objects under varying
  illumination conditions}, in: \bibinfo{booktitle}{2003 IEEE Computer Society
  Conference on Computer Vision and Pattern Recognition, 2003. Proceedings.},
  \bibinfo{organization}{IEEE}. pp. \bibinfo{pages}{I--I}.
\bibitem[{Hong et~al.(2006)Hong, Wright, Huang and Ma}]{hong2006multiscale}
\bibinfo{author}{Hong, W.}, \bibinfo{author}{Wright, J.},
  \bibinfo{author}{Huang, K.}, \bibinfo{author}{Ma, Y.}, \bibinfo{year}{2006}.
\newblock \bibinfo{title}{Multiscale hybrid linear models for lossy image
  representation}.
\newblock \bibinfo{journal}{IEEE Transactions on Image Processing}
  \bibinfo{volume}{15}, \bibinfo{pages}{3655--3671}.
\bibitem[{Hu et~al.(2016)Hu, Short, Riggan, Gordon, Gurton, Thielke, Gurram and
  Chan}]{hu2016polarimetric}
\bibinfo{author}{Hu, S.}, \bibinfo{author}{Short, N.J.},
  \bibinfo{author}{Riggan, B.S.}, \bibinfo{author}{Gordon, C.},
  \bibinfo{author}{Gurton, K.P.}, \bibinfo{author}{Thielke, M.},
  \bibinfo{author}{Gurram, P.}, \bibinfo{author}{Chan, A.L.},
  \bibinfo{year}{2016}.
\newblock \bibinfo{title}{A polarimetric thermal database for face recognition
  research}, in: \bibinfo{booktitle}{Proceedings of the IEEE conference on
  computer vision and pattern recognition workshops}, pp.
  \bibinfo{pages}{187--194}.
\bibitem[{Hunter and Strohmer(2010)}]{hunter2010performance}
\bibinfo{author}{Hunter, B.}, \bibinfo{author}{Strohmer, T.},
  \bibinfo{year}{2010}.
\newblock \bibinfo{title}{Performance analysis of spectral clustering on
  compressed, incomplete and inaccurate measurements}.
\newblock \bibinfo{journal}{arXiv preprint arXiv:1011.0997} .
\bibitem[{Ji et~al.(2017)Ji, Zhang, Li, Salzmann and Reid}]{ji2017deep}
\bibinfo{author}{Ji, P.}, \bibinfo{author}{Zhang, T.}, \bibinfo{author}{Li,
  H.}, \bibinfo{author}{Salzmann, M.}, \bibinfo{author}{Reid, I.},
  \bibinfo{year}{2017}.
\newblock \bibinfo{title}{Deep subspace clustering networks}, in:
  \bibinfo{booktitle}{Advances in Neural Information Processing Systems}, pp.
  \bibinfo{pages}{24--33}.
\bibitem[{Kingma and Ba(2014)}]{kingma2014adam}
\bibinfo{author}{Kingma, D.P.}, \bibinfo{author}{Ba, J.}, \bibinfo{year}{2014}.
\newblock \bibinfo{title}{Adam: A method for stochastic optimization}.
\newblock \bibinfo{journal}{arXiv preprint arXiv:1412.6980} .
\bibitem[{Korona and Kokar(1996)}]{korona1996model}
\bibinfo{author}{Korona, Z.}, \bibinfo{author}{Kokar, M.M.},
  \bibinfo{year}{1996}.
\newblock \bibinfo{title}{Model theory based fusion framework with application
  to multisensor target recognition}, in: \bibinfo{booktitle}{1996
  IEEE/SICE/RSJ International Conference on Multisensor Fusion and Integration
  for Intelligent Systems (Cat. No. 96TH8242)}, \bibinfo{organization}{IEEE}.
  pp. \bibinfo{pages}{9--16}.
\bibitem[{Lee et~al.(2005)Lee, Ho and Kriegman}]{lee2005acquiring}
\bibinfo{author}{Lee, K.C.}, \bibinfo{author}{Ho, J.},
  \bibinfo{author}{Kriegman, D.J.}, \bibinfo{year}{2005}.
\newblock \bibinfo{title}{Acquiring linear subspaces for face recognition under
  variable lighting}.
\newblock \bibinfo{journal}{IEEE Transactions on pattern analysis and machine
  intelligence} \bibinfo{volume}{27}, \bibinfo{pages}{684--698}.
\bibitem[{Li and Vidal(2015)}]{li2015structured}
\bibinfo{author}{Li, C.G.}, \bibinfo{author}{Vidal, R.}, \bibinfo{year}{2015}.
\newblock \bibinfo{title}{Structured sparse subspace clustering: A unified
  optimization framework}, in: \bibinfo{booktitle}{Proceedings of the IEEE
  conference on computer vision and pattern recognition}, pp.
  \bibinfo{pages}{277--286}.
\bibitem[{Lin et~al.(2011)Lin, Liu and Su}]{lin2011linearized}
\bibinfo{author}{Lin, Z.}, \bibinfo{author}{Liu, R.}, \bibinfo{author}{Su, Z.},
  \bibinfo{year}{2011}.
\newblock \bibinfo{title}{Linearized alternating direction method with adaptive
  penalty for low-rank representation}, in: \bibinfo{booktitle}{Advances in
  neural information processing systems}, pp. \bibinfo{pages}{612--620}.
\bibitem[{Liu et~al.(2012)Liu, Lin, Yan, Sun, Yu and Ma}]{liu2012robust}
\bibinfo{author}{Liu, G.}, \bibinfo{author}{Lin, Z.}, \bibinfo{author}{Yan,
  S.}, \bibinfo{author}{Sun, J.}, \bibinfo{author}{Yu, Y.},
  \bibinfo{author}{Ma, Y.}, \bibinfo{year}{2012}.
\newblock \bibinfo{title}{Robust recovery of subspace structures by low-rank
  representation}.
\newblock \bibinfo{journal}{IEEE transactions on pattern analysis and machine
  intelligence} \bibinfo{volume}{35}, \bibinfo{pages}{171--184}.
\bibitem[{Luenberger et~al.(1984)Luenberger, Ye et~al.}]{luenberger1984linear}
\bibinfo{author}{Luenberger, D.G.}, \bibinfo{author}{Ye, Y.}, et~al.,
  \bibinfo{year}{1984}.
\newblock \bibinfo{title}{Linear and nonlinear programming}.
  volume~\bibinfo{volume}{2}.
\newblock \bibinfo{publisher}{Springer}.
\bibitem[{Ng et~al.(2002)Ng, Jordan and Weiss}]{ng2002spectral}
\bibinfo{author}{Ng, A.Y.}, \bibinfo{author}{Jordan, M.I.},
  \bibinfo{author}{Weiss, Y.}, \bibinfo{year}{2002}.
\newblock \bibinfo{title}{On spectral clustering: Analysis and an algorithm},
  in: \bibinfo{booktitle}{Advances in neural information processing systems},
  pp. \bibinfo{pages}{849--856}.
\bibitem[{Ngiam et~al.(2011)Ngiam, Khosla, Kim, Nam, Lee and
  Ng}]{ngiam2011multimodal}
\bibinfo{author}{Ngiam, J.}, \bibinfo{author}{Khosla, A.},
  \bibinfo{author}{Kim, M.}, \bibinfo{author}{Nam, J.}, \bibinfo{author}{Lee,
  H.}, \bibinfo{author}{Ng, A.}, \bibinfo{year}{2011}.
\newblock \bibinfo{title}{Multimodal deep learning (pp. 689--696)}, in:
  \bibinfo{booktitle}{International conference on machine learning (ICML),
  Bellevue, WA}.
\bibitem[{Ramachandram and Taylor(2017)}]{ramachandram2017deep}
\bibinfo{author}{Ramachandram, D.}, \bibinfo{author}{Taylor, G.W.},
  \bibinfo{year}{2017}.
\newblock \bibinfo{title}{Deep multimodal learning: A survey on recent advances
  and trends}.
\newblock \bibinfo{journal}{IEEE Signal Processing Magazine}
  \bibinfo{volume}{34}, \bibinfo{pages}{96--108}.
\bibitem[{Rockafellar(1974)}]{rockafellar1974augmented}
\bibinfo{author}{Rockafellar, R.T.}, \bibinfo{year}{1974}.
\newblock \bibinfo{title}{Augmented lagrange multiplier functions and duality
  in nonconvex programming}.
\newblock \bibinfo{journal}{SIAM Journal on Control} \bibinfo{volume}{12},
  \bibinfo{pages}{268--285}.
\bibitem[{Roheda et~al.(2018a)Roheda, Krim, Luo and Wu}]{roheda2018decision}
\bibinfo{author}{Roheda, S.}, \bibinfo{author}{Krim, H.}, \bibinfo{author}{Luo,
  Z.Q.}, \bibinfo{author}{Wu, T.}, \bibinfo{year}{2018}a.
\newblock \bibinfo{title}{Decision level fusion: An event driven approach}, in:
  \bibinfo{booktitle}{2018 26th European Signal Processing Conference
  (EUSIPCO)}, \bibinfo{organization}{IEEE}. pp. \bibinfo{pages}{2598--2602}.
\bibitem[{Roheda et~al.(2019)Roheda, Krim, Luo and Wu}]{roheda2019event}
\bibinfo{author}{Roheda, S.}, \bibinfo{author}{Krim, H.}, \bibinfo{author}{Luo,
  Z.Q.}, \bibinfo{author}{Wu, T.}, \bibinfo{year}{2019}.
\newblock \bibinfo{title}{Event driven fusion}.
\newblock \bibinfo{journal}{arXiv preprint arXiv:1904.11520} .
\bibitem[{Roheda et~al.(2020a)Roheda, Krim and Riggan}]{roheda2020commuting}
\bibinfo{author}{Roheda, S.}, \bibinfo{author}{Krim, H.},
  \bibinfo{author}{Riggan, B.S.}, \bibinfo{year}{2020}a.
\newblock \bibinfo{title}{Commuting conditional gans for multi-modal fusion},
  in: \bibinfo{booktitle}{ICASSP 2020-2020 IEEE International Conference on
  Acoustics, Speech and Signal Processing (ICASSP)},
  \bibinfo{organization}{IEEE}. pp. \bibinfo{pages}{3197--3201}.
\bibitem[{Roheda et~al.(2020b)Roheda, Krim and Riggan}]{roheda2020robust}
\bibinfo{author}{Roheda, S.}, \bibinfo{author}{Krim, H.},
  \bibinfo{author}{Riggan, B.S.}, \bibinfo{year}{2020}b.
\newblock \bibinfo{title}{Robust multi-modal sensor fusion: An adversarial
  approach}.
\newblock \bibinfo{journal}{IEEE Sensors Journal} \bibinfo{volume}{21},
  \bibinfo{pages}{1885--1896}.
\bibitem[{Roheda et~al.(2018b)Roheda, Riggan, Krim and Dai}]{roheda2018cross}
\bibinfo{author}{Roheda, S.}, \bibinfo{author}{Riggan, B.S.},
  \bibinfo{author}{Krim, H.}, \bibinfo{author}{Dai, L.}, \bibinfo{year}{2018}b.
\newblock \bibinfo{title}{Cross-modality distillation: A case for conditional
  generative adversarial networks}, in: \bibinfo{booktitle}{2018 IEEE
  International Conference on Acoustics, Speech and Signal Processing
  (ICASSP)}, \bibinfo{organization}{IEEE}. pp. \bibinfo{pages}{2926--2930}.
\bibitem[{Soong and Rosenberg(1988)}]{soong1988use}
\bibinfo{author}{Soong, F.K.}, \bibinfo{author}{Rosenberg, A.E.},
  \bibinfo{year}{1988}.
\newblock \bibinfo{title}{On the use of instantaneous and transitional spectral
  information in speaker recognition}.
\newblock \bibinfo{journal}{IEEE Transactions on Acoustics, Speech, and Signal
  Processing} \bibinfo{volume}{36}, \bibinfo{pages}{871--879}.
\bibitem[{Taylor et~al.(2016)Taylor, Shai, Stanley and
  Mucha}]{taylor2016enhanced}
\bibinfo{author}{Taylor, D.}, \bibinfo{author}{Shai, S.},
  \bibinfo{author}{Stanley, N.}, \bibinfo{author}{Mucha, P.J.},
  \bibinfo{year}{2016}.
\newblock \bibinfo{title}{Enhanced detectability of community structure in
  multilayer networks through layer aggregation}.
\newblock \bibinfo{journal}{Physical review letters} \bibinfo{volume}{116},
  \bibinfo{pages}{228301}.
\bibitem[{Valada et~al.(2016)Valada, Oliveira, Brox and
  Burgard}]{valada2016deep}
\bibinfo{author}{Valada, A.}, \bibinfo{author}{Oliveira, G.L.},
  \bibinfo{author}{Brox, T.}, \bibinfo{author}{Burgard, W.},
  \bibinfo{year}{2016}.
\newblock \bibinfo{title}{Deep multispectral semantic scene understanding of
  forested environments using multimodal fusion}, in:
  \bibinfo{booktitle}{International Symposium on Experimental Robotics},
  \bibinfo{organization}{Springer}. pp. \bibinfo{pages}{465--477}.
\bibitem[{Wang et~al.(2018)Wang, Skau, Krim and Cervone}]{wang2018fusing}
\bibinfo{author}{Wang, H.}, \bibinfo{author}{Skau, E.}, \bibinfo{author}{Krim,
  H.}, \bibinfo{author}{Cervone, G.}, \bibinfo{year}{2018}.
\newblock \bibinfo{title}{Fusing heterogeneous data: A case for remote sensing
  and social media}.
\newblock \bibinfo{journal}{IEEE Transactions on Geoscience and Remote Sensing}
  \bibinfo{volume}{56}, \bibinfo{pages}{6956--6968}.
\bibitem[{Xu et~al.(1992)Xu, Krzyzak and Suen}]{xu1992methods}
\bibinfo{author}{Xu, L.}, \bibinfo{author}{Krzyzak, A.}, \bibinfo{author}{Suen,
  C.Y.}, \bibinfo{year}{1992}.
\newblock \bibinfo{title}{Methods of combining multiple classifiers and their
  applications to handwriting recognition}.
\newblock \bibinfo{journal}{IEEE transactions on systems, man, and cybernetics}
  \bibinfo{volume}{22}, \bibinfo{pages}{418--435}.
\bibitem[{Yang et~al.(2006)Yang, Rao and Ma}]{yang2006robust}
\bibinfo{author}{Yang, A.Y.}, \bibinfo{author}{Rao, S.R.}, \bibinfo{author}{Ma,
  Y.}, \bibinfo{year}{2006}.
\newblock \bibinfo{title}{Robust statistical estimation and segmentation of
  multiple subspaces}, in: \bibinfo{booktitle}{2006 Conference on Computer
  Vision and Pattern Recognition Workshop (CVPRW'06)},
  \bibinfo{organization}{IEEE}. pp. \bibinfo{pages}{99--99}.
\bibitem[{Zhu et~al.(2019)Zhu, Hui, Zhang, Du, Wen and Hu}]{zhu2019multi}
\bibinfo{author}{Zhu, P.}, \bibinfo{author}{Hui, B.}, \bibinfo{author}{Zhang,
  C.}, \bibinfo{author}{Du, D.}, \bibinfo{author}{Wen, L.},
  \bibinfo{author}{Hu, Q.}, \bibinfo{year}{2019}.
\newblock \bibinfo{title}{Multi-view deep subspace clustering networks}.
\newblock \bibinfo{journal}{arXiv preprint arXiv:1908.01978} .

\end{thebibliography}

	
	\bio{}
	$\bf{Sally \ Ghanem}$ received the B.Sc. degree in electrical engineering from Alexandria University, in 2013, and the M.Sc. degree in electrical and computer engineering from North Carolina State University, Raleigh, NC, USA in 2016, where she is currently pursuing the Ph.D. degree. During the PhD program, she has spent time at Oak Ridge National Laboratory working on multimodal data. Her research interests include the areas of computer vision, digital signal processing, image processing and machine learning.
	\endbio
	
	\bio{}
	$\bf{Hamid \ Krim}$ received the B.Sc. and M.Sc. and Ph.D. in ECE. He was a Member of Technical Staff at AT\&T Bell Labs, where he has conducted research and development in the areas of telephony and digital communication systems/subsystems. Following an NSF Postdoctoral Fellowship at Foreign Centers of Excellence, LSS/University of Orsay, Paris, France, he joined the Laboratory for Information and Decision Systems, MIT, Cambridge, MA, USA, as a Research Scientist and where he performed/supervised research. He is currently a Professor of electrical engineering in the Department of Electrical and Computer Engineering, North Carolina State University, NC, leading the Vision, Information, and Statistical Signal Theories and Applications Group. His research interests include statistical signal and image analysis and mathematical modeling with a keen emphasis on applied problems in classification and recognition using geometric and topological tools. He has served on the SP society Editorial Board and on TCs, and is the SP Distinguished Lecturer for 2015-2016.
	\endbio

\end{document}